\documentclass[10pt, final, letter, twocolumns]{IEEEtran}

\usepackage{wrapfig}
\usepackage{amsmath, amssymb, amsfonts}
\usepackage{latexsym}
\usepackage{alltt}
\usepackage{graphicx}
\usepackage{url}
\usepackage{color, colortbl}
\usepackage{float} 
\usepackage{multicol}
\usepackage{multirow}
\usepackage{hhline}
\usepackage{graphicx}
\usepackage{booktabs}
\usepackage{rotating}
\usepackage{algorithm}
\usepackage{algorithmicx}
\usepackage[caption=false,font=footnotesize]{subfig}
\usepackage{algpseudocode}
\usepackage{tabto}
\usepackage{lipsum}
\usepackage{siunitx}
\usepackage[flushleft]{threeparttable}
\usepackage{cite}

\newcommand{\exclude}[1]{}

\definecolor{Gray}{gray}{0.9}
\definecolor{LightCyan}{rgb}{0.88,1,1}

\include{psfig}

\usepackage[normalem]{ulem}

\begin{document}

\title{Design Rule Violation Hotspot Prediction Based on Neural Network Ensembles}
\author{
\IEEEauthorblockN{Wei Zeng, Azadeh Davoodi, and Yu Hen Hu}

\IEEEauthorblockA{University of Wisconsin--Madison\\
\{wei.zeng, adavoodi, yhhu\}@wisc.edu}
}

\maketitle
\bstctlcite{IEEEexample:BSTcontrol}
\begin{abstract}
Design rule check is a critical step in the physical design of integrated circuits to ensure  manufacturability. However, it can be done only after a time-consuming detailed routing procedure, which adds drastically to the time of design iterations. With advanced technology nodes, the outcomes of global routing and detailed routing become less correlated, which adds to the difficulty of predicting design rule violations from earlier stages. In this paper, a framework based on neural network ensembles is proposed to predict design rule violation hotspots using information from placement and global routing. A soft voting structure and a PCA-based subset selection scheme are developed on top of a baseline neural network from a recent work. Experimental results show that the proposed architecture achieves significant improvement in model performance compared to the baseline case. For half of test cases, the performance is even better than random forest, a commonly-used ensemble learning model.
\end{abstract}

\vspace{-2mm}
\section{Introduction}\vspace{-1mm}
\label{sec:intro}
Today's IC fabrication technologies
require satisfying many complex design rules to ensure manufacturability. Creating a layout that is clean of design rule violations has turned into a cumbersome task, requiring many iterations in the design flow, and consequently impacting the time-to-market. 

Within the design flow, Design Rule Check (DRC) is typically applied after  detailed routing. However, the process of  detailed routing can be rather tedious
and expensive, which typically takes several hours, if not days,
to finish. Therefore, it is highly  desirable if an inexpensive DRC 
predictor is developed so that DRC hotspot locations on the layout may be predicted
accurately at the earlier stages in the design flow. In this
way, a designer may leverage this early feedback without going through  detailed routing and DRC phases each time.

Recent research has focused on prediction of DRC\ hotspots, and on placement and/or global routing techniques to minimize the violations after detailed routing \cite{VLSIDAT17, DAC18, ISPD17}. They have  identified  various features at the placement and/or global routing stages which can contribute to DRC violations. The   process involves extracting these features during placement and/or global routing  followed by machine learning for modeling. 

A significant challenge during modeling is effective use of a large number of extracted features. Direct incorporation of all the features during modeling can cause overfitting issues, besides significant increase in the time to do the modeling itself. To handle these challenges,  Chan et al.\  proposed different schemes to define  smaller subsets of  features and conducted a study to find the most useful subset  \cite{ISPD17}. These subsets however were defined in an empirical / non-systematic manner.

Recently, Tabrizi et al.\ used a Neural Network (NN) model for DRC hotspot prediction \cite{DAC18}, where the network model was composed of only a single hidden layer. 

\textbf{In this paper}, we apply NN ensembles \cite{NNE90, NNE95, NNE02zhou} on top of the baseline model in \cite{DAC18} to improve DRC hotspot prediction. 


To further improve the performance in NN ensembles, we propose a feature selection scheme based on principal component analysis (PCA) \cite{PCA}. Related works on feature selection in NN ensembles include \cite{FS92ga, FS99opitz, FS00, FS10}, which select features by using different training samples and/or are based on an objective function of the features that evaluates their fitness. Our method differs from these works in that we randomly select the features in PCA-transformed linear subspace, which does not have any objective function to evaluate (which saves the training time), and does not require different training samples for each learner (which makes it easier to be embedded on the original NN).

To summarize, our contributions are listed below.
\begin{itemize}
\item Our   workflow receives as input a wide range of features inspired by a comprehensive study of DRC  prediction. The feature size is more than any single prior work. 
\item We offer systematic techniques to generate useful feature subsets for use in our model by applying PCA, subset selection, and a smart random selection scheme.
\item These are incorporated within a proposed NN ensemble architecture that can be easily implemented with popular NN frameworks like Tensorflow \cite{tensorflow} or Keras \cite{keras}.
\end{itemize}

In our experiments conducted using the ISPD 2015 detailed routability-driven placement benchmarks \cite{ISPD15benchmark}, we consider 387 features and show significant improvement compared to a baseline architecture inspired by \cite{DAC18}.
We also obtain improvement in over half of the designs compared to random forest \cite{random-forest}, a commonly-used ensemble learning model.

The rest of this paper is organized as follows. Section~\ref{sec:overview} introduces the overall workflow of DRC hotspot prediction. Section~\ref{sec:feature} presents a study on features used in related works, and how we extract features and ground truths in our problem. Section~\ref{sec:model} elaborates the proposed model and Section~\ref{sec:model-train-valid-eval} shows how it is used. Section~\ref{sec:results} gives the experimental results.
\vspace{-3mm}
\section{DRC Hotspot Prediction Workflow}
\label{sec:overview}

\begin{figure}
\centering
\includegraphics[width=0.46\textwidth]{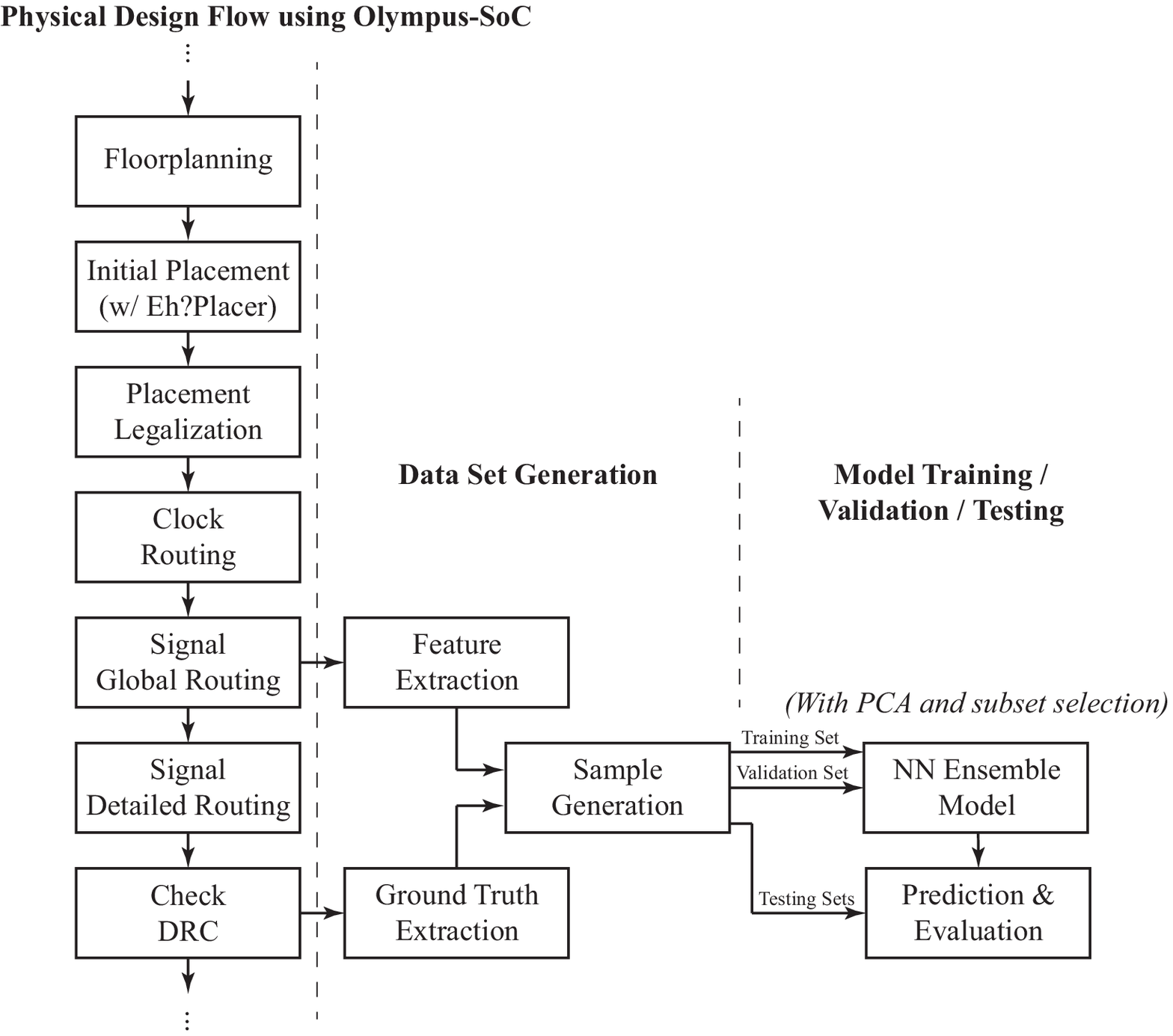}\vspace{-6mm}
\caption{DRC hotspot prediction workflow.}
\label{fig:flow}\vspace{-7mm}
\end{figure}

\exclude{
The left panel in Fig.~\ref{fig:flow} shows the physical design flow. In global routing for signal nets, the layout is divided into grids known as global routing cells (g-cells). According to the net connections in the design, the input/output of standard cells need to be connected with a set of wires in one or more metal layers. If a wire connects standard cells in different g-cells, it must go across at least one g-cell borders. If a wire encompasses more than one metal layer, a via will also be needed to connect the wire from one layer to another. Global routing places the wires and vias in the granularity of g-cells, and makes an initial estimation on the routing resources (i.e., wires across g-cell borders and vias inside the g-cell) needed and allowed for each g-cell. Then the detailed router will be invoked to assign the actual locations and shapes for the wires within the g-cell, where to across the border, and where to place the vias within the g-cell, etc., followed by a series of procedures trying to fix the routing and design rule violations. Afterward, a design rule checking (DRC) module will be invoked to uncover any design rule violations that cannot be fixed by the router. We refer to a g-cell with at least one DRC violations after detailed routing as a \emph{DRC hotspot}. With this information, designers can either manually fix the violations if feasible, or initiate another routing process with different settings in these regions, in hopes that the violations will be reduced.

The process of signal detailed routing can be rather tedious and expensive, which typically takes several hours, if not days, to finish, while a global routing only takes several minutes. Therefore, it would be desirable if an inexpensive DRC hotspot predictor is developed so that DRC hotspots may be predicted accurately once the signal global routing phase is done. In this way, designs may leverage this early feedback in the design iterations without going through the detailed routing and DRC checking phases each time.


The DRC hotspot prediction problem can be stated as follows: Given the signal global routing outcome, predict whether a g-cell, after signal detailed routing and DRC checking, will be deemed a DRC hotspot. The prediction is made based on routed designs with the same technology and same physical design flow. 
}
The DRC hotspot prediction problem is stated as follows: Given a 
global routing outcome, predict whether a g-cell, after detailed routing, will contain any DRC violation. The prediction is made based on other routed areas/designs with the same technology and same physical design flow.

We show the overall workflow in
Fig.~\ref{fig:flow}. Our approach is to formulate this problem as a supervised classification problem. We extract features from placement and global routing   (shown in the left panel of Fig.~\ref{fig:flow}) to form a feature vector for each g-cell, and then develop an NN ensemble model that accept this feature vector as input and produce a binary output indicating whether the corresponding g-cell may be a DRC hotspot. 

The data gathering process is shown in the middle panel in Fig.~\ref{fig:flow}. We use 14 designs (including two hidden designs) with a 65\,nm technology from the ISPD 2015 contest benchmark suite \cite{ISPD15benchmark}\footnote{Design \texttt{edit\_dist\_a} is excluded from our experiments since it took more than 10 days to route and is therefore considered unroutable. \texttt{superblue} designs are also excluded because the technology is different.}.  They are listed in Table~\ref{tab:dataset}. Each design is first fed into \texttt{Eh?Placer} \cite{ehplacer}, which produces a placed .def file. Then with Olympus-SoC \cite{olympus}, we do the following steps:
(1) Placement legalization, (2) Global and detailed routing for the clock net, (3) Global routing for signal nets,  (4) Detailed routing for signal nets, and (5)\ Check for DRC\ violations. 
 After step (3), the intermediate results will be sent to a feature extraction module to extract the feature vectors, and the DRC errors in step (5) will be used to produce the ground truth. The methods will be discussed in Section~\ref{sec:feature}.


These data samples are then partitioned into a training set, a validation set, and several test sets. We use the training set to train our proposed voting-based NN model, and use the validation set to tune the hyperparameters with grid search. The trained model is then applied to test sets to predict the DRC hotspots and evaluate the performance. 


For each design, we randomly split the g-cells by the ratios of 20\%, 20\% and 60\% as training, validation and testing samples, respectively. There are two exceptions: a) g-cells that fully overlap with a macro are excluded from the data sets before splitting, since empirically no routed wire and via can exist in such g-cells; b) all data in designs \texttt{fft\_b}, \texttt{matrix\_mult\_2} and \texttt{matrix\_mult\_c} will be allocated into the test sets that are unforeseen in training and validation phases, which allows us to better examine the generalization performance of the proposed model. We combine all training samples into a training set, and all validation samples into a validation set, so that both of them contain a mixture of g-cells from different designs with the same technology file. We build 14 test sets, each containing testing samples in one design of the benchmarks suite, so that we can observe the model performance for different designs. Note that the data samples are highly imbalanced. According to Table~\ref{tab:dataset}, only 2616 out of 146090 (1.8\%) g-cells are DRC hotspots, i.e. positive samples.

\begin{table}[t]
\caption{The Profile of Designs}\vspace{-3mm}
\label{tab:dataset}
\centering
\tabcolsep=3.0pt
\scriptsize
\begin{tabular}{lrrrrr}
\toprule
\multirow{2}{*}{Design} & \multirow{2}{*}{\# G-cells} & \multirow{2}{*}{\# DRC hotspots} & Training & Validation & Testing\\
 &&& samples &  samples &  samples\\
\midrule
des\_perf\_1 & 5476 & 676 & 1095 & 1095 & 3286\\
des\_perf\_a & 11498 & 246 & 2300 & 2300 & 6898\\
des\_perf\_b & 10000 & 0 & 2000 & 2000 & 6000\\
fft\_1 & 1936 & 50 & 387 & 387 & 1162\\
fft\_2 & 3249 & 17 & 650 & 650 & 1949 \\
fft\_a & 6491 & 2 & 1298 & 1298 & 3895\\
fft\_b & 6506 & 534 & 0 & 0 & 6506\\
matrix\_mult\_1 & 8281 & 154 & 1656 & 1656 & 4969\\
matrix\_mult\_2 & 8464 & 193 & 0 & 0 & 8464 \\
matrix\_mult\_a & 21757 & 13 & 4351 & 4351 & 13055\\
matrix\_mult\_b & 24257 & 613 & 4851 & 4851 & 14555\\
matrix\_mult\_c & 24213 & 62 & 0 & 0 & 24213 \\
pci\_bridge32\_a & 3569 & 56 & 714 & 714 & 2141\\
pci\_bridge32\_b & 10393 & 0 & 2079 & 2079 & 6235\\
\midrule
Total & 146090 & 2616 & 21381 & 21381 & 103328\\
\bottomrule
\end{tabular}\vspace{-3mm}
\end{table}


\section{Feature and Ground Truth Extraction}
\label{sec:feature}

\begin{figure}
\centering
\includegraphics[height=4cm]{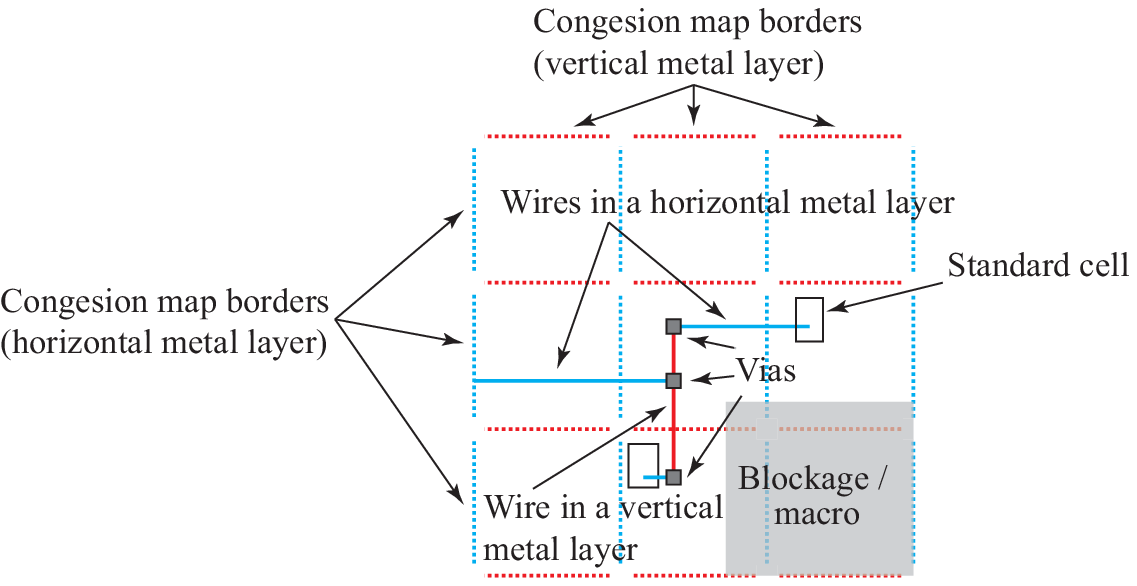}\vspace{-2mm}
\caption{A 3$\times$3 g-cell window with standard cells,
wires (different colors indicate different metal layers), vias, congestion map borders,  blockage/macro. }\vspace{-5mm}
\label{fig:layout-illustration}
\end{figure}

\exclude{
\begin{figure}
\centering
\hfill
\subfloat[]{\includegraphics[width=0.2\textwidth]{figs/cong-map-edges-hori.eps}}
\hfill
\subfloat[]{\includegraphics[width=0.2\textwidth]{figs/cong-map-edges-vert.eps}}
\hfill
\caption{A $3\times3$ window in (a) a horizontal metal layer, and (b) a vertical metal layer. Congestion map edges are marked in bold solid segments.}
\label{fig:cong-map-edges}
\end{figure}
}
\exclude{
In Fig.~\ref{fig:layout-illustration}, we plot an illustrative window with g-cells (with dashed lines), standard cells, wires across g-cell borders, vias inside the g-cells, and a blockage/macro where no wires or vias are allowed in general. In prior works, a variety of aspects are considered as related to routability and thus the number of DRC violations, including but not limited to:
\begin{itemize}
\item Location of g-cells \cite{VLSIDAT17, DAC18}, with the observation that g-cells at the center are more likely to be congested.
\item Density-related information, based on the heuristic that a higher density of cells and/or pins often means more difficulty to route:
\begin{itemize}
\item cell density (in terms of either number or percentage) \cite{VLSIDAT17, DAC18, ISPD15, ISPD17, VLSIDAT18},
\item pin density (in terms of either number or percentage) \cite{VLSIDAT17, ISPD15, ISPD17, VLSIDAT18, ASQED15, ICCD16, DAC12},
\item pin spacing / distribution (various definitions) \cite{DAC18, ISPD17, ICCD16}.
\end{itemize}
\item Special pins and cells, which may have special (usually stricter) constraints in routing: 
\begin{itemize}
\item clock pins \cite{DAC18},
\item pins in nets with non-default rules (NDRs) \cite{DAC18, ISPD15},
\item multi-height and sequential cells \cite{ISPD17},
\item cells with a higher DRC error rate than average \cite{ISPD17}.
\end{itemize}
\item Connectivity, where complex connections around the g-cells may lead to more difficulty in routing:
\begin{itemize}
\item local nets \cite{ISPD15, ISPD17, VLSIDAT18, ASQED15, ICCD16},
\item local net wirelength \cite{ASQED15, DAC12, ISPD15},
\item cross-border nets and its pins \cite{DAC18, ICCD16, ISPD17, VLSIDAT18},
\item degrees of nets connected to each pin \cite{VLSIDAT17}.
\end{itemize}
\item Congestion map capacity, load and overflow \cite{DAC18,  ISPD15, ISPD17, VLSIDAT18, ASQED15}, which are direct indicators of supply and demand of routing resources.
\item Blockages for placement and/or routing \cite{VLSIDAT17, DAC18, ASQED15}, which further limit the routing resources.
\end{itemize}
Recent works \cite{VLSIDAT17, DAC18, ISPD17, VLSIDAT18} also extract features in a window including neighboring g-cells to consider their contributions to the routability.
}

\vspace{-1mm}
\label{sec:prior}

Fig. \ref{fig:layout-illustration} illustrates the types of layout information available at the placement and global routing stages. 
Using these, prior works have defined various features which relate to routability and thus the DRC hotspots prediction, including: 
\begin{itemize}\vspace{-1mm}
\item Location of g-cells in the layout\cite{VLSIDAT17, DAC18},
\item Density-related information (e.g., cell density  \cite{VLSIDAT17,
DAC18, ISPD15, ISPD17, VLSIDAT18},
pin density  \cite{VLSIDAT17,
ISPD15, ISPD17, VLSIDAT18, DAC12, ASQED15, ICCD16},
 pin spacing / distribution  \cite{DAC18, ISPD17,
ICCD16}),
\item Special pins and cells, which may have  constraints
in routing (e.g., clock pins \cite{DAC18}, pins in nets with non-default
rules (NDRs) \cite{DAC18, ISPD15}, and multi-height cells \cite{ISPD17}),
\item Connectivity, where complex connections around the g-cells may complicate routing (e.g.,
local nets \cite{ISPD15, ISPD17, VLSIDAT18, ASQED15, ICCD16},
 and cross-border nets  \cite{DAC18, ICCD16, ISPD17, VLSIDAT18}),
\item Congestion map, load, overflow \cite{DAC18,  ISPD15, ISPD17,
VLSIDAT18, ASQED15}, which indicate supply and demand of
routing resources,
\item  Blockages for placement and/or routing \cite{VLSIDAT17, DAC18, ASQED15},
which further limit the routing resources.\vspace{-1mm}
\end{itemize}
Recent works \cite{VLSIDAT17, DAC18, ISPD17, VLSIDAT18} also extract features
in a window including neighboring g-cells to consider their contributions.

These observations motivate us to extract the following features
in the designs, from the placed cells and the congestion
map after signal global routing which can be explained using Fig. \ref{fig:layout-illustration}. Each sample corresponds to a g-cell in the layout, which is expanded to a $3\times3$ \textit{window} consisting of this g-cell (referred to as ``central g-cell'') and its 8 neighbors.
\begin{itemize}
\item For each of 9 g-cells in the window,

\begin{itemize}
\item The center $x$- and $y$-coordinates, normalized to $[0,1]$ by dividing the $x$- (or $y$-) coordinate of the g-cell center by the layout width (or height).
\item The number of standard cells fully within the g-cell.
\item The number of pins fully within the g-cell.
\item The number of clock pins fully within the g-cell.
\item The number of local nets, defined as nets whose all pins are within the same g-cell.
\item The number of pins that belong to any local net.
\item The number of pins that have NDRs, as defined in the ISPD 2015 contest benchmarks in our experiments.
\item The pin spacing, defined as the arithmetic mean of pair-wise\footnote{We have $n\choose 2$ such pairs for a g-cell with $n$ pins.} Manhattan distances of pins inside g-cell.
\item The percentage of area occupied by blockages.
\item The percentage of area occupied by standard cells.
\end{itemize}
\item For each of 12 congestion border edges (i.e., segments with blue/red dots  in Fig.~\ref{fig:layout-illustration}) in \emph{each metal layer}, and for each of 9 g-cells inside the window in \emph{each via layer},
\begin{itemize}
\item The capacity $C$, defined as the maximum allowed number of wires/vias across the edge. 
\item The load $L$, defined as the number of wires that are already across the edge (for metal layers) / the number of vias inside the g-cell (for via layers).
\item The difference of $C$ and $L$.
\end{itemize}
\exclude{
\item 
\begin{itemize}
\item The capacity $C$, defined as the maximum allowed number of vias inside the g-cell. 
\item The load $L$, defined as the number of vias that are already occupied inside the g-cell.
\item The difference of $C$ and $L$.
\end{itemize}
}
\end{itemize}

We include almost all features from prior works\footnote{Some
features such as the number of multi-height cells were not applicable
because they did not exist in our benchmark suite.}. This results in a large
number of features, 387 in total.

For the ground truth, we examine the bounding boxes of DRC errors as reported by Olympus-SoC. A g-cell is a DRC hotspot if and only if the g-cell overlaps with any DRC error bounding box. A sample is positive if and only if the central g-cell is a DRC hotspot.



\vspace*{-3pt}
\section{Proposed Neural Network Ensemble Model}
\begin{figure}
\centering
\includegraphics[width=0.47\textwidth]{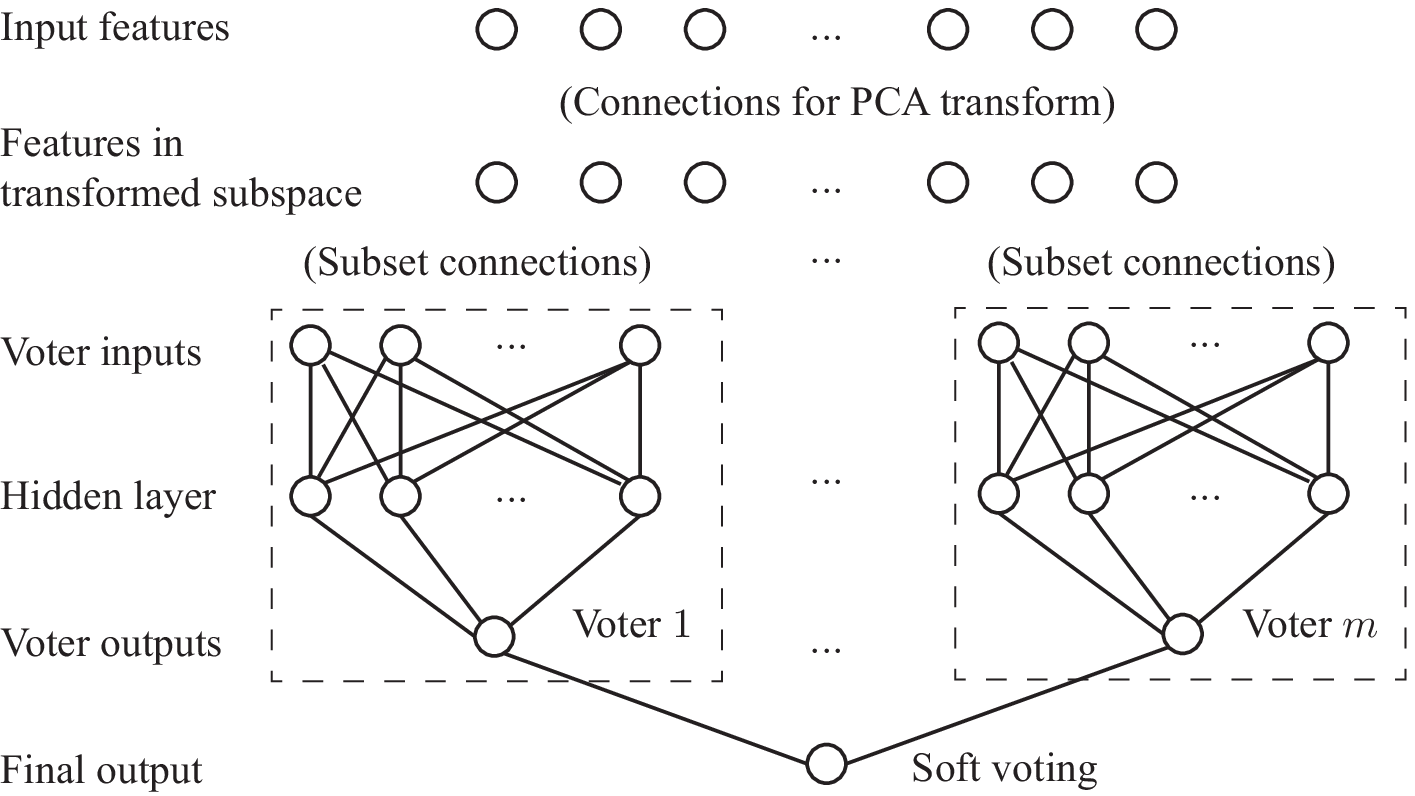}\vspace{-3mm}
\caption{The proposed NN ensemble model with PCA transform, subset selection, and soft voting embedded.}\label{fig:voter}\vspace{-6mm}
\end{figure}

\vspace{-1mm}
\label{sec:model}
In this section we present the proposed NN ensemble model as illustrated in Fig.~\ref{fig:voter}. It contains a PCA transform layer and subset connections for feature selection, a group of voters that will be trained to classify the samples, and an arbitration structure to combine the outputs of each voter. The network inputs are the extracted features, each normalized to zero mean and unit variance for better numerical robustness and training convergence. The inputs of each voter are a subset of the network inputs. The subsets for each voter may be the same or different as per the setting. Note that even if some voters use the same input subset, their outputs can still be different due to random initialization of network weights for each voter. Next, we show each model component in detail. 
\vspace*{-3pt}
\subsection{The Baseline Neural Network}\vspace{-1mm}
\label{subsec:baseline}
In a recent work \cite{DAC18}, a simple NN with one hidden layer of 20 neurons is adopted to predict the short violations after detailed routing from placed netlists. Since \cite{DAC18} does not provide more details about the architecture, we assume no dropout or regularization is applied, and we use ReLU as the activation function of the hidden layer, and the sigmoid function for the only neuron in the output layer. Formally, the activation of the $i$-th neuron in the hidden layer is given by
\vspace{-2mm}
\begin{equation}
a_{i}^{(1)}=ReLU\left(\sum_{j=1}^r w_{ij}^{(1)}x_{j}+b_{i}^{(1)}\right),~i=1,2,\dots,20,
\label{eqn:hidden}
\end{equation}
where $r$ is the number of neurons of the input layer, $x_{j}$ is the $j$-th (normalized) feature, and $w_{ji}^{(1)}$ and $b_{i}^{(1)}$ are weights and biases of the $i$-th neuron in the hidden layer, respectively.
The activation of the output neuron is given by
\vspace{-2mm}\begin{equation}
a^{(2)}=\sigma\left(\sum_{i=1}^{20} w_{i}^{(2)}a_{i}^{(1)}+b^{(2)}\right),\\
\end{equation}
where $w_{i}^{(2)}$ and $b^{(2)}$ are weights and the bias of the output neuron, respectively. In these two equations,
$ReLU(\cdot)$ and $\sigma(\cdot)$ are ReLU and sigmoid functions, respectively.

This network is used as voters in Fig.~\ref{fig:voter}. It also serves as the baseline model to illustrate how our proposed NN ensembles can improve the performance for DRC hotspot prediction.

\vspace{-1mm}
\subsection{Voters and Soft Voting}\vspace{-1mm}
\label{subsec:soft-voting}
According to the research in ensemble learning \cite{NNE90, NNE95}, an ensemble model has a better overall performance than individual learners if the following conditions are satisfied: (1) Learners are different when making decisions, and
(2) Each individual learner performs better than random guessing. 

Inspired by this property, we develop a voting structure, where each voter is the same baseline NN as described in Section~\ref{subsec:baseline}. The inputs of each voter are a subset of features in PCA transformed subspace (details are described in subsections below). Note that in practice, even if all voters use the same subset as inputs, their outputs are prone to be different after training due to random initialization of network weights for the voters, which naturally satisfy the first condition above.

We use \emph{soft voting} to arbitrate from all voter outputs. In this way, each voter output is considered as a \emph{probability} that the sample is positive,  and the final output takes the sum of voter outputs. Soft voting is easy to implement in corporation of NN, as the summation is essentially a fully-connected layer with all weights of $1$ and a bias of $0$. It is also more flexible because the final output is a continuous variable and the user can apply different thresholds of classification to get different prediction results, according to various factors (e.g., the costs of having false positive and false negative samples).

\subsection{PCA Transform and Subset Connection Layers}\vspace{-1mm}
\label{subsec:PCA}
In this work, we select a subset of features for each voter. Intuitively, manual feature selection seems unnecessary in an NN, as it can ``learn'' new features itself in the hidden layers. However, too many input features in a network may contribute to model overfitting due to the added network complexity. Therefore it helps if the inputs can be pre-processed with a smaller size before flowing into the voters. Another motivation for feature selection in NN ensembles is to promote the diversity among voters, which can help with performance improvement.

According to \eqref{eqn:hidden}, each neuron in the hidden layer (before activation) is a \emph{linear combination} of input features. Therefore, it is theoretically equivalent to express the inputs in any bases that span the same linear subspace of input features. This motivates us to apply PCA, which involves a linear transform, on the input features and use the features in the transformed subspace. The reason is that, if some input features are collinear (or highly correlated), there will be transformed features with zero (or very small) variances, which can be discarded with no or limited loss on the input data. By doing so, the network complexity can be reduced, which relieves the overfitting and thus improves its performance.

To this end, we first apply PCA on the (normalized\footnote{We use normalized features before PCA so that the effect of different feature means and variances is eliminated. The resulting PCA transformation matrix is therefore solely dependent on the correlations among features.}) input features using samples from the training set, and store the transformation matrix as weights in the PCA transform layer in Fig.~\ref{fig:voter}, so that the transform will be applied to all data samples in training, validation and test sets before flowing downstream. In the rest of the paper, we will use \emph{transformed features}, or simply features, to refer to features in PCA transformed subspace, unless otherwise specified.

After PCA, we select a subset of transformed features to reduce the network complexity. In the context of NN, one natural way is to connect each voter input neuron directly to one of transformed features. For convenience and flexibility, however, we implement a subset connection layer as the input layer of each voter. It is a fully-connected layer with weights of either $1$ or $0$ and biases of all zeros. For example, if we have $5$ transformed features, each voter has $3$ input neurons, and the first three features are selected as subset for Voter $1$, then the weights and biases for the inputs of Voter $1$ are
$$\mathbf{W}_1=\left[\begin{matrix}
 1 & 0 & 0 & 0 & 0 \\
 0 & 1 & 0 & 0 & 0 \\
 0 & 0 & 1 & 0 & 0 \\
 \end{matrix}\right], \mathbf{b}_1 = \left[\begin{matrix}
 0 \\
 0 \\
 0 \\
 \end{matrix}\right],$$
respectively. These weights and biases are assigned before training and do not change during training.
\subsection{Smart Random Selection of PCA Transformed Features}\vspace{-1mm}
\label{subsec:smartPC}
A conventional way of feature selection with PCA is to keep a group of transformed features with highest variances. Although it should work in most cases, it has two problems in the setting of classification with a voting network. First, a large variance of PCA transformed features do not always translate to good ability for classification, as PCA does not consider the class labels in the training set. Second, each voter will get the same subset of features, so the overall  improvement (compared to fewer or just one voters) may be limited.

To address these problems, we propose Smart Random Selection (SRS), which aims to choose different yet good subsets of features for each voter. It is based on the heuristic similar to that of PCA: (transformed) features with larger variances are generally more valuable for classification and are therefore more encouraged to be included in the subset. With SRS, the transformed features are randomly selected one-at-a-time until the number of selected features is fulfilled. In each iteration, a feature is selected with the probability proportional to its variance. In this way, features with larger variances are more likely to be included in the subset, while voters can still get different subsets of features. The steps of SRS is shown as Algorithm~\ref{alg:smart}, where $n$ is the subset size, $N$ is the total number of features, and $var[i]$ is the variance of the $i$-th feature.
\vspace{-2mm}
\begin{algorithm}[H]
\caption{\textsc{SmartRandom}\,$(n, var[1\dots N])$}
\label{alg:smart}
\begin{algorithmic}[1]
\State $S \leftarrow \varnothing$, $T \leftarrow \{1,2,\dots,N\}$.
\While {$|S| < n$}
\For {$i \in T$}
\State $p_i=var[i]/\sum_{j\in T} var[j]$.
\EndFor
\State Randomly select a feature $k\in T$ with probability $p_k$. 
\State $S\leftarrow S\cup{\{k\}}$, $T \leftarrow \{1,2,\dots,N\}\setminus S$.
\EndWhile
\end{algorithmic}
\end{algorithm}
\vspace{-2mm}

We should note that, although SRS uses a similar heuristic to that of PCA, the aforementioned problems of PCA can both be relieved. First, SRS introduces randomness in feature selection, so that each voter can get a different subset of features. Second, owing to the randomness, most features have a good chance to be included in the subset of at least one voter (unless it has zero variance which means it is actually redundant), which reduces the odds of discarding a good feature by mistake. Our experiments show that SRS actually outperforms conventional dimension reduction with PCA.

\vspace{-1mm}
\subsection{Comparison of the Proposed Model and Random Forest}\vspace{-1mm}
The proposed NN ensemble model shares some similarities to random forest (RF), a famous ensemble model for supervised classification. Both methods improve the model performance by combining a group of learners, and each individual learner uses different features. The main difference of these two models lies on the learners and the way to diversify them. The learners in the proposed model are NNs, which are more flexible and easier to train with back-propagation, while RF uses decision trees, which are simpler and more interpretable. Also, RF diversifies individual learners by using random feature subsets and random training samples for each learner, while the proposed model use SRS to select different subsets of PCA transformed features for each voter. According to our experiments (not shown here due to page limit), using random training samples on top of the proposed model does not show extra performance improvements.

\vspace{-2mm}
\section{Model Training, Validation and Evaluation}
\vspace{-1mm}
\label{sec:model-train-valid-eval}
During training, only the weights and biases inside voters (i.e., ``hidden layer'' and ``voter output'' in Fig.~\ref{fig:voter}) are trainable. Layers with preset weights and biases (i.e., PCA transform, subset connection and voting) are fixed. To compute the training loss, we compare each \emph{voter output} with the same ground truth from the training set. The loss function is defined as the sum of cross-entropy losses w.r.t.\ each voter outputs. To address the imbalance of data sets, we also assign different weights for positive and negative samples in the loss function to address the imbalance of data. Formally, with $m$ voters and $n$ samples in the training set, the training loss
\vspace{-2mm}\begin{equation}
L = \sum_{i=1}^m \sum_{j=1}^n -w_1y_j\log p_{ij} - w_0(1-y_j)\log(1-p_{ij}),
\end{equation}
where $p_{ij}$ is the output (i.e., sigmoid activation) of voter $i$ for sample $j$, $y_j$ is the ground truth of sample $j$, $w_0$ and $w_1$ are the weights for negative and positive samples, respectively.

To validate and tune the hyperparameters (including learning rate, number of training epochs, number of voters, subset size) of the model, we train a series of models with different hyperparameters, feed the validation set into each trained model, and compare the model performance (see below). We choose the model that performs best with validation set as final, and feed it with test sets for prediction and evaluation.

Due to the imbalance of positive and negative samples in the data set, accuracy (i.e., the percentage of correctly predicted samples) is not a good indicator of model performance. Consider an extreme case where there are only $1\%$ positive samples. A model could have an accuracy of $99\%$ even if it ignores all features and predicts every sample as negative. To address this, the following metrics are often seen in literature.
\begin{itemize}
\item $TPR$: true positive rate, a.k.a. sensitivity or recall,
\item $TNR$: true negative rate, a.k.a. specificity,
\item $FPR$: false positive rate $=1-TNR$,
\item $FNR$: false negative rate $=1-TPR$,
\item $Prec$: precision.
\end{itemize}
Although these metrics are better alternatives to accuracy, all of them can change when different thresholds of classification are applied. In practice, as mentioned in Section~\ref{subsec:soft-voting}, the user is free to adjust the threshold to get different prediction results with the same model. With this consideration, the receiver operating characteristic (ROC) curve (i.e., TPR-FPR curve) and precision-recall curve (P-R curve) are better indicators of overall model quality, since they show the model performance at essentially every threshold. We also need a metric similar to accuracy, yet independent of the threshold.
Therefore, we use the following metrics for model validation and evaluation in this paper. All of them are threshold-independent.  

\begin{itemize}\vspace{-1mm}
\item $Acc_{e}$: effective accuracy, defined as $TPR$ at the threshold of classification such that $TPR=TNR$,
\item $A_{roc}$: area under ROC curve,
\item $A_{prc}$: area under P-R curve.
\end{itemize}\vspace{-1mm}

\vspace{-1mm}
\section{Simulation Results}\vspace{-1mm}
\label{sec:results}
We run numerical experiments in a Linux workstation with an Intel 6-core 2.93\,GHz CPU, an Nvidia GTX 1080Ti GPU, and 24\,GB memory. The proposed model was implemented using Keras with Tensorflow backend and GPU accelaration. The inputs of the model are the 387 normalized features described in Section~\ref{sec:feature} (before PCA transform). We train the model with the following hyperparameters. The optimizer for minimizing the loss is Adam \cite{adam} with learning rate $0.001$. The number of training epochs is $50$. The class weight is $1:10$, meaning that the loss of a positive sample is counted as $10$ times of that of a negative sample. These hyperparameters are chosen based on the best result in the validation set. 

\vspace{-1mm}
\subsection{Improvements by Soft Voting, PCA transform and SRS}\vspace{-1mm}
To show the efficacy of soft voting, PCA transform and SRS as described in Section~\ref{sec:model}, we show experimental results with the settings shown in Table~\ref{tab:settings}. Note that setting~1 is essentially the baseline model presented in Section~\ref{subsec:baseline}.

\begin{table}
\scriptsize
\centering
\caption{Model Settings Used in Experiments}\vspace{-3mm}
\label{tab:settings}
\begin{tabular}{rrrcc}
\toprule
ID & \# voters & \# inputs/voter & PCA transform & Subset selection\\
\midrule
1  & 1 & 387 & No & No\\
2 & 100 & 387 & No & No\\
3 & 100 & 20 & Yes & Largest variance\\
4 & 100 & 20 & Yes & SRS\\
\bottomrule
\end{tabular}\vspace{-7mm}
\end{table}

We show in Table~\ref{tab:results} the model performance measured with these four settings. The bold font indicates the best performance among these settings. Note that the results for \texttt{des\_perf\_b} and \texttt{pci\_bridge32\_b} are undefined, because there is no positive sample in these data sets so that some of the underlying metrics (i.e., $TPR$ and $Prec$) are undefined.

\begin{table*}
\scriptsize
\centering
\caption{Experiment Results with Different Settings}\vspace{-3mm}
\label{tab:results}
\begin{tabular}{l|rrr|rrr|rrr|rrr||rrr}
\toprule
Setting & \multicolumn{3}{c|}{1 (Baseline)} & \multicolumn{3}{c|}{2} & \multicolumn{3}{c|}{3} & \multicolumn{3}{c||}{4 (Proposed)} & \multicolumn{3}{c}{Random forest (RF)}\\
Test set & $Acc_{e}$ & $A_{roc}$ & $A_{prc}$ & $Acc_{e}$ & $A_{roc}$ & $A_{prc}$ & $Acc_{e}$ & $A_{roc}$ & $A_{prc}$ & $Acc_{e}$ & $A_{roc}$ & $A_{prc}$ & $Acc_{e}$ & $A_{roc}$ & $A_{prc}$ \\
\midrule
perf\_1 & 0.9166&       0.9687& 0.7961& 0.9313& 0.9798& 0.8718  &0.9363&        0.9799& 0.8745  &{\bf 0.9393} &{\bf 0.9810}   &{\bf 0.8894} & {\bf 0.9424}& {\bf 0.9827}&   {\bf 0.8924}\\
perf\_a & 0.8838&       0.9636& 0.6445& 0.8903& 0.9687& 0.6719& {\bf 0.9553}    &0.9897&        0.8072& 0.9550  &{\bf 0.9913}&        {\bf 0.8328} & {\bf 0.9594}  &0.9881 &       0.8109\\
fft\_1 & 0.8930 &       0.9565  &0.7215 &0.9001 &0.9674 &0.7217 &0.9045 &0.9837 &0.7773 &{\bf 0.9278}&        {\bf 0.9885}    &{\bf 0.8387} & {\bf 0.9696}   & {\bf 0.9972}  & {\bf 0.9030}\\
fft\_2 & 0.9930 &0.9921 &0.2121 &0.9959 &0.9949 &0.2851 &0.9954 &0.9957 &0.4218 &{\bf 0.9961} &{\bf 0.9963}   &{\bf 0.4677} & 0.9956 & {\bf 0.9981}  & {\bf 0.5892}\\
fft\_a & 0.8013 &0.6027 &0.0003 &{\bf 0.8463} & {\bf 0.6926}    &{\bf 0.0004} &       0.7976 &        0.5953  &0.0003 &0.7571 &       0.5141 &        0.0003 & 0.4177 &       0.4177& 0.0001\\
fft\_b & 0.6835&        0.7185& 0.2003& 0.6798& 0.7713& 0.2286& {\bf 0.7375}&   {\bf 0.8336}&        {\bf 0.3598} &  0.7282& 0.8282  &0.3475 & 0.7217 & 0.7811 &      {\bf 0.3585}\\
mult\_1 & 0.8775        & 0.9401 &      0.3283  &0.9191&        0.9753  &0.4745 &0.9207 &0.9695&        0.4145& {\bf 0.9380}&        {\bf 0.9812}    &{\bf 0.5429} & {\bf 0.9407}  & 0.9793        & 0.4246\\
mult\_2 & 0.8043        &0.8967 &0.2277 &0.8556 &0.9345 &0.3568 &0.8393 &0.9290 &0.2705 &{\bf 0.8808} &{\bf 0.9515}   &{\bf 0.4365} & {\bf 0.9083} & {\bf 0.9607} & 0.4213\\
mult\_a & 0.9968&       0.9983  &0.6052 &0.9980 &0.9983 &0.1363 &{\bf 0.9998}   &{\bf 0.9998} &{\bf 0.6869}   &0.9997 &{\bf 0.9998}   &0.5740 & 0.9966        &0.9979&        0.5289\\
mult\_b &0.9309 &0.9818 &0.6664 &0.9418 &0.9873 &0.7409 &{\bf 0.9473}   &{\bf 0.9893} &0.7670 &0.9454 &0.9889 &{\bf 0.7719} & 0.9553        &0.9868 &0.7689\\
mult\_c & 0.9546        &0.9814&        0.1019& 0.9539& 0.9829& 0.1162& 0.9661& 0.9910&{\bf         0.2113}&        {\bf 0.9677} &  {\bf 0.9914} &  0.2034 & 0.9662        &0.9854 & {\bf 0.3568}\\
b32\_a & 0.8905 &  0.9495 &        0.6443 &        0.9461 &        {\bf 0.9882} &       0.7816 &        {\bf 0.9473} &  0.9825 &        0.8299 &        0.9411 &       0.9834  & {\bf 0.8486} & {\bf 0.9506} &    0.9769 &        0.7142\\
\midrule
All testing & \multirow{2}{*}{0.8364}   & \multirow{2}{*}{0.9113} &     \multirow{2}{*}{0.4462} &       \multirow{2}{*}{0.8554} &       \multirow{2}{*}{0.9407} & \multirow{2}{*}{0.5260} &       \multirow{2}{*}{0.8955} &       \multirow{2}{*}{0.9654} &       \multirow{2}{*}{0.5729} &       \multirow{2}{*}{\bf 0.9003} &   \multirow{2}{*}{\bf 0.9682} &   \multirow{2}{*}{\bf 0.6082} & \multirow{2}{*}{\bf 0.9041} & \multirow{2}{*}{0.9508} &  \multirow{2}{*}{\bf 0.6125}\\
samples &  &&&&&&&&&&&&&&\\
\bottomrule
\end{tabular}\vspace{-7mm}
\end{table*}

Several observations can be made from Table~\ref{tab:results}.
\begin{itemize}\vspace{-1mm}
\item Comparing setting~1 (i.e., the baseline model) and setting 2, where the number of voters increases from 1 to 100, the model performance improves for all test sets owing to the soft voting as described in Section~\ref{subsec:soft-voting}.
\item Comparing settings~2 and 3, where we use raw features verses 20 most variant PCA transformed features, all but one test set show improvement of the model performance introduced by PCA transform and subset selection as described in Section~\ref{subsec:PCA}.
\item Comparing settings~3 and 4, where the PCA transformed features are selected in the conventional way verses SRS, most test sets show improvement of the model performance introduced by SRS as described in Section~\ref{subsec:smartPC}.
\end{itemize}
\vspace{-1mm}

\begin{figure}
\centering
\hfill
\subfloat[]{\includegraphics[width=0.24\textwidth]{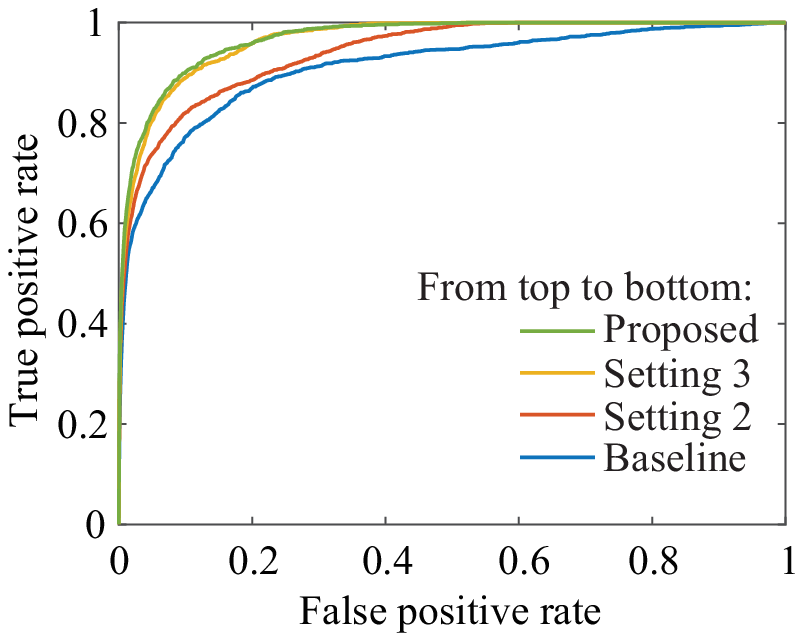}}
\hfill
\subfloat[]{\includegraphics[width=0.24\textwidth]{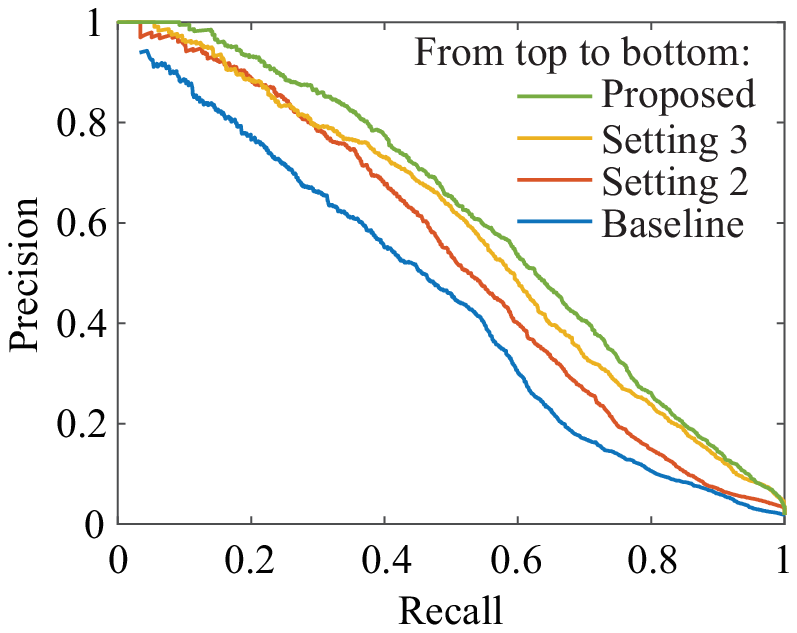}}
\hfill
\vspace{-2mm}
\caption{(a) ROC curve and (b) precision-recall curve of NN ensembles with different settings (all testing samples included).}
\vspace{-4mm}
\label{fig:roc-prc-curves}
\end{figure}

Similarly, by comparing the last row of Table~\ref{tab:results}, where all testing samples are taken into consideration, we can confirm the improvements of overall performance by virtue of each component in our model introduced in Section~\ref{sec:model}. 
Fig.~\ref{fig:roc-prc-curves} shows the corresponding ROC curve and precision-recall curve, where the performance improvements introduced by each component of the model can be clearly observed as the gaps between lines.

Model training (with setting~4) takes up to $3$ hours with our hardware devices, and prediction takes less than a minute per test set, much less than the time required for detailed routing.

\vspace{-2mm}
\subsection{NN Ensemble vs. RF}\vspace{-1mm}
For comparison, we also list the performance with RF in Table~\ref{tab:results}, which is configured as having 100 voters, each using at most 20 (raw) features. The bold font indicates that RF performs better than the proposed NN ensemble (setting~4).
\exclude{
\begin{table}
\scriptsize
\centering
\caption{Experiment Results with Random Forest}\vspace{-3mm}
\label{tab:rf}
\begin{tabular}{l|rrr}
\toprule
Setting & \multicolumn{3}{c}{Random Forest}\\
Test set & $ACC_{eff}$ & $A_{ROC}$ & $A_{PRC}$ \\
\midrule
perf\_1 & {\bf 0.9424}& {\bf 0.9827}&   {\bf 0.8924}\\
perf\_a & {\bf 0.9594}  &0.9881 &       0.8109\\
perf\_b & --- & --- & ---\\
fft\_1 & {\bf 0.9696}   & {\bf 0.9972}  & {\bf 0.9030}\\
fft\_2 & 0.9956 & {\bf 0.9981}  & {\bf 0.5892}\\
fft\_a & 0.4177 &       0.4177& 0.0001\\
fft\_b & 0.7217 & 0.7811 &      {\bf 0.3585}\\
mult\_1 & {\bf 0.9407}  & 0.9793        & 0.4246\\
mult\_2 & {\bf 0.9083}  &{\bf 0.9607}   &0.4213\\
mult\_a & 0.9966        &0.9979&        0.5289\\
mult\_b & 0.9553        &0.9868 &0.7689\\
mult\_c & 0.9662        &0.9854 & {\bf 0.3568}\\
bridge32\_a & {\bf 0.9506} &    0.9769 &        0.7142
\\
bridge32\_b &  --- & --- & ---\\
\midrule
All testing samples &  \multirow{1}{*}{\bf 0.9041} & \multirow{1}{*}{0.9508} &  \multirow{1}{*}{\bf 0.6125}\\
\bottomrule
\end{tabular}\vspace{-8mm}
\end{table}
}
Comparing the columns of ``setting~4'' and ``RF'' in Table~\ref{tab:results}, we can see the proposed NN ensemble model achieves similar, if not better, performance to that of RF in terms of effective accuracy, areas under ROC curve and precision-recall curve. At least half of test sets shows better performance in terms of any metric (e.g., 8 out of 12 test sets have better $A_{roc}$) using the proposed model than RF. Therefore, the proposed model can be used as a good complement to RF.
\exclude{

\begin{table}[t]
\centering
\caption{Information on the  CNN models and ReRAM\ architectures of LeNet300-100 for techniques 1 \& 2.}
\tabcolsep=3.0pt
\scriptsize
\label{tab:lnet300-models}
\begin{tabular}{l|c|c|c|c}
\toprule
\textbf{Layer}                     & 
\textbf{\begin{tabular}[c]{@{}c@{}}Weight Matrix \\ In$\times$Out\end{tabular}} & \textbf{ReRAM$^+$ / ReRAM$^-$} & \textbf{\begin{tabular}[c]{@{}c@{}}\#DAC\\/split\end{tabular}} & \textbf{\begin{tabular}[c]{@{}c@{}}\#ADC\\/split\end{tabular}}  \\
\midrule
\rowcolor{LightCyan}\multicolumn{5}{c}{\textbf{Base Architecture}}                                                                                \\ \midrule
\textbf{FC1}                       & 784$\times$300               & (512$\times$300) (272$\times$300)      & 784                                                            & 1200                                                             \\
\textbf{FC2}                       & 300$\times$100               & (300$\times$100)                & 300                                                            & 200                                                             \\
\textbf{FC3}                       & 100$\times$10                & (100$\times$10)                 & 100                                                            & 20                                                              \\ \hline
\textbf{Total}                     &                         &                          & 1184                                                           & 1420                                                             \\ \midrule
\rowcolor{LightCyan}\multicolumn{5}{c}{\textbf{Architecture 1: Inserting  layer FC3' of dimension (100$\times$100)}}                                                                                                                                                      \\ \midrule
\textbf{FC1}                       & 784$\times$300             & (512$\times$300) (272$\times$300)      & 784                                                            & 1200                                                             \\ 
\textbf{FC2}                       & 300$\times$100             & (300$\times$100)                & 300                                                            & 200                                                              \\ \rowcolor{Gray}
\textbf{FC3'}                      & 100$\times$100             & (100 $\times$100)                & 100                                                            & 200                                                              \\
\textbf{FC3}                       & 100$\times$10              & (100$\times$10)                 & 100                                                            & 20                                                              \\ \hline
\textbf{Total}                     &                         &                          & 1284                                                           & 1620                                                             \\ \midrule
\rowcolor{LightCyan}\multicolumn{5}{c}{\textbf{Architecture 2: Adding 75 neurons to FC2}}\\    
\midrule                                                         
{\textbf{FC1}}   & 784$\times$300             & (512$\times$300) (272$\times$300)      & 784                                                            & 1200  \\ \rowcolor{Gray}
{\textbf{FC2}}   & 300$\times$175             & (300$\times$175)                & 300                                                            & 350                              \\ \rowcolor{Gray}
{\textbf{FC3}}   & 175$\times$100               & (175$\times$100)                & 175                                                            & 200                                                     \\ \hline
{\textbf{Total}} &                         &                          & 1259                                                           & 1750                                                       \\ \bottomrule
\end{tabular}
\end{table}
}

\vspace{-1mm}
\section{Conclusions}
\label{sec:conclusions}
In this paper, we use NN ensembles to predict DRC hotspot in early stages of physical design. With a systematic feature subset selection scheme, the performance of the proposed model is significantly improved over a single NN, and is better or comparable to that of RF. The proposed NN ensemble model can be easily implemented with popular NN frameworks with affordable computational cost.

\scriptsize
\bibliographystyle{IEEEtran}


\end{document}